\documentclass[sigconf]{acmart}
\AtBeginDocument{%
  }

\copyrightyear{2026}
\acmYear{2026}
\setcopyright{cc}
\setcctype{by}
\acmConference[MM '26]{Proceedings of the 34th ACM International Conference on Multimedia}{November 10--14, 2026}{Rio de Janeiro, Brazil}
\acmBooktitle{Proceedings of the 34th ACM International Conference on Multimedia (MM '26), November 10--14, 2026, Rio de Janeiro, Brazil}
\acmDOI{10.1145/3767308.3835932}
\acmISBN{979-8-4007-2213-4/2026/11}

\usepackage{balance}
\usepackage{multirow}
\usepackage{makecell}
\begin{document}

\title[Geometry-Aware Test-Time Learning for Quantitative Spatial Reasoning]{Geometry-Aware Test-Time Learning for \\ Quantitative Spatial Reasoning}


\author{Gege Zhang}
\authornote{Equal contribution $^\dagger$Corresponding author}
\affiliation{%
  \institution{South China University of Technology}
  \city{Guangzhou}
  \country{China}
}
\email{ftzgg@mail.scut.edu.cn}

\author{Shuaicheng Niu}
\authornotemark[1]
\affiliation{%
  \institution{Nanyang Technological University}
  \city{Singapore}
  \country{Singapore}
}
\email{shuaicheng.niu@ntu.edu.sg}

\author{Gang Dai}
\affiliation{%
  \institution{Guangdong University of Technology}
  \city{Guangzhou}
  \country{China}
}
\email{daigang@gdut.edu.cn}

\author{Lei Sun}
\affiliation{%
  \institution{South China University of Technology}
  \city{Guangzhou}
  \country{China}
}
\email{sunlei@guangzhou.csg.cn}

\author{Shuangping Huang}
\authornotemark[2]
\affiliation{
  \institution{South China University of Technology}
  \city{Guangzhou}
  \country{China}
}
\email{eehsp@scut.edu.cn}






\renewcommand{\shortauthors}{Gege Zhang et al.}

\begin{abstract}
    Quantitative spatial reasoning in visual-language models (VLMs) aims to infer spatial distances and directional relationships among objects in 3D space from a 2D image and a natural language query. Despite recent progress, VLM spatial reasoning remains brittle under distribution shifts, largely due to the high cost of 3D supervision. As a result, models often produce inconsistent or contradictory predictions when faced with novel object configurations or rephrased spatial queries, revealing a misalignment between learned representations and underlying geometry. To address this, we propose \textbf{TTL-SR}, a geometry-aware \textbf{T}est-\textbf{T}ime \textbf{L}earning framework for quantitative \textbf{S}patial \textbf{R}easoning that leverages geometric consistency constraints and unlabeled test data to adapt models to target domains. 
    Specifically, TTL-SR augments the input query with geometrically coupled auxiliary queries, filters unreliable predictions via adaptive geometric triggering to construct structured token-level pseudo-labels, and updates model parameters under a geometry-aware multi-objective loss using only test data.
    Experimental results demonstrate that TTL-SR significantly boosts spatial reasoning performance, yielding 6.47\% and 9.41\% accuracy gains for Qwen3-VL-4B-Instruct and SpatialRGPT-VILA-1.5-8B on Q-Spatial-ScanNet dataset, respectively. 
    
\end{abstract}




\maketitle

\section{Introduction}
\begin{figure}[t]
\centerline
{
\includegraphics[width=1.0\columnwidth]{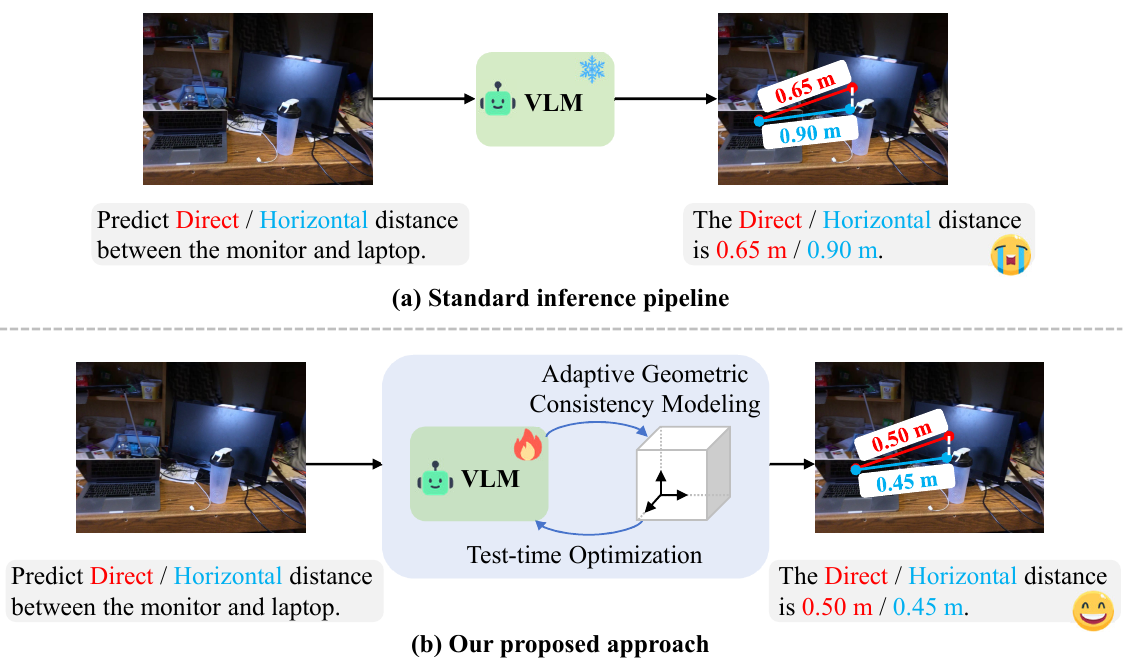}
}
    \caption{Comparison of spatial reasoning paradigms. (a) Standard VLM inference is static, often leading to inconsistent and physically implausible predictions. (b) Our proposed approach leverages geometric consistency at test time, enabling coherent and physically plausible spatial reasoning.}
    \label{fig:intro}
\end{figure}
Quantitative spatial reasoning in visual-language models (VLMs) estimates 3D distances and directions from a single image and language query, supporting embodied AI \cite{azzolini2025cosmos}, robotic manipulation \citep{shridhar2022cliport, pan2025omnimanip}, and augmented reality \cite{elford2022exploring}. 
Despite recent VLM progress, this task remains a challenging problem, as it requires precise modeling of geometric relationships and spatial structure.

Recent studies improve VLM spatial reasoning mainly through fine-tuning on synthesized spatial VQA data \citep{ma2024spatialpin, mouselinos-etal-2024-beyond, rizvi2024sparc, chen2024spatialvlm, ranasinghe2024learning, cheng2024spatialrgpt, cai2025spatialbot}. However, costly 3D annotations restrict training coverage, leaving spatial knowledge implicitly encoded in fixed parameters. At test time, this knowledge is not explicitly activated or jointly constrained, so models treat inter-dependent geometric attributes in isolation and often produce physically inconsistent predictions under novel scenes, as shown in Figure~\ref{fig:intro} (a).


To address the above issue, a natural next step is to introduce learning objectives at test time that allow the model to adapt to the test data distribution. Although explicit spatial annotations are unavailable during inference, the model has access to abundant unlabeled test inputs, making Test-Time Learning (TTL) a practical solution. TTL updates model parameters during inference and has demonstrated improved robustness and generalization in vision recognition models~\cite{wang2021tent, fleuret2021test, zhang2024come} and large language models across diverse reasoning tasks \citep{zuo2025ttrl, pmlr-v267-hu25z, acikgoz2025self}.

However, existing TTL methods are primarily designed for standard vision tasks (e.g., classification and segmentation) and typically rely on confidence-based objectives, such as entropy minimization~\cite{wang2021tent, pmlr-v267-hu25z}. When directly applied to quantitative spatial reasoning, these objectives often prove ineffective or even harmful. Intuitively, while confidence maximization encourages compact and homogeneous representations that benefit classification tasks, as in Figure~\ref{distance_acc}, quantitative spatial reasoning requires preserving fine-grained geometric distinctions to support accurate continuous predictions. As a result, naive confidence optimization tends to over-collapse representations and disrupt the underlying geometric structure of the feature space. This limitation highlights the need for TTL objectives that go beyond prediction confidence and explicitly enforce geometric consistency during inference.

In this paper, we propose \textbf{TTL-SR}, a geometry-aware \textbf{T}est-\textbf{T}ime \textbf{L}earning framework for enhanced quantitative \textbf{S}patial \textbf{R}easoning in VLMs. The key challenge we address lies in bridging explicit physical geometry with the inherently discrete and probabilistic output space of language models. Instead of treating spatial reasoning as isolated predictions, TTL-SR grounds TTL in fundamental geometric constraints, e.g., metric relations implied by the Pythagorean theorem, which must be jointly satisfied across multiple spatial queries. As shown in Figure~\ref{fig:intro}, to operationalize these constraints during inference, we design a query augmentation strategy that systematically probes inter-dependent geometric attributes under the same visual scene. The resulting predictions, expressed as token-level probability distributions over the language model vocabulary, are then aligned and transformed into a differentiable, self-supervised learning objective. This geometry-consistent objective enables reliable parameter updates at test time. As a result, TTL-SR actively corrects physically implausible predictions and enforces global geometric coherence, leading to substantially improved performance for quantitative spatial reasoning.

\textbf{Main novelty and contributions:}
\textbf{1)} We propose \textbf{TTL-SR}, a model-agnostic, plug-and-play TTL framework that integrates with pre-trained VLMs to enhance quantitative spatial reasoning.
\textbf{2)} We design a geometry-consistent test-time objective based on structured spatial query augmentations, which aligns discrete language model predictions with continuous geometric relations and supports stable, self-supervised parameter updates.
\textbf{3)} Extensive experiments demonstrate that TTL-SR consistently improves accuracy, robustness, and generalization across spatial reasoning benchmarks on both specialized and general-purpose VLMs.

\section{Related Work}
\subsection{Spatial Reasoning in VLMs}
Spatial reasoning has become a key challenge in vision-language research. Early studies improve spatial understanding using additional sensory cues, such as depth \cite{fichtl2014learning} or sequential RGB observations \cite{migimatsu2022grounding}. In contrast, our work focuses on the more challenging single-image setting.
Recent advances primarily enhance spatial reasoning through training-stage improvements. A dominant line of work fine-tunes VLMs on curated spatial VQA datasets to strengthen spatial associations and numerical prediction capabilities \cite{chen2024spatialvlm, cai2025spatialbot, wu2024mind, ranasinghe2024learning}. Complementarily, SpatialReasoner-R1 \cite{shen2025fine} adopts fine-grained preference optimization to elicit Chain-of-Thought reasoning patterns for solving geometric problems. Other approaches incorporate external geometric priors, such as zero-shot frameworks that leverage 3D foundation models \cite{ma2024spatialpin}, or region-aware architectures that improve grounding through explicit spatial localization \cite{guo2024regiongpt, cheng2024spatialrgpt, yuan2024osprey}. These methods improve spatial reasoning by either enriching supervision signals or introducing inductive biases into model architectures.
Related vision research has also explored structure-aware generation, visual in-context guidance, cross-modal alignment, and metric learning, which provide broader context for modeling visual structure and correspondence~\cite{dai2023disentangling, dai2024one, dai2025beyond, dai2025vg, peng2026dual, peng2025proxy, peng2025globally}.
In parallel, several benchmarks have been proposed to evaluate spatial reasoning capabilities, including Q-Spatial Bench \cite{liao-etal-2024-reasoning}, SpatialRGPT-Bench \cite{cheng2024spatialrgpt}, Spatial-MM \cite{shiri2024empirical}, VSI-Bench \cite{yang2025thinking}, MindCube\cite{yin2025spatial}, SPAR-Bench \cite{zhang2025from}, and ViewSpatial-Bench \cite{li2025viewspatial}, which cover diverse spatial reasoning tasks and evaluation protocols.

Nevertheless, existing methods typically acquire spatial knowledge from a narrow range of training scenarios due to the high cost of 3D supervision. Since spatial knowledge is injected through training-time parameter updates, it remains implicitly embedded in fixed model parameters and cannot be activated at test time.

\subsection{Test-Time Learning for Reasoning}
TTL has emerged as a compelling paradigm that enables models to adapt during inference using only unlabeled test data. In vision, early Test-Time Adaptation (TTA) methods such as Tent~\cite{wang2021tent} established entropy minimization as a simple unsupervised update rule, followed by works that improve stability, calibration, and anti-collapse behavior through uncertainty calibration, sharpness and feature regularization, self-bootstrapping, or asymmetric entropy optimization~\cite{tan2025uncertainty, niu2026adapt, pmlr-v267-niu25a, chen2026zerosiam}. 
Beyond vision recognition, test-time learning has recently been extended to reasoning models. Reflexion~\cite{shinn2023reflexion} improves test-time reasoning through iterative self-feedback, while TTRL~\cite{zuo2025ttrl} formulates test-time adaptation as reinforcement learning. TLM~\cite{pmlr-v267-hu25z} performs self-supervised adaptation via perplexity minimization with sample selection, EMPO~\cite{zhang2025right} proposes unsupervised incentivization, and \cite{wei2025first} studies unsupervised post-training for multi-modal large language models. Test-time optimization has also been explored in diffusion models through guided trajectory refinement~\cite{dai2026guided}.

Although existing TTL approaches enhance general reasoning, they primarily rely on self-certainty-driven objectives. These methods lack the mechanisms to model structured numerical or physical constraints, which are essential for quantitative spatial reasoning. Consequently, they remain prone to producing logically consistent but physically impossible results, necessitating a TTL paradigm explicitly tailored to quantitative spatial reasoning.

\begin{figure*}[t]
    \centerline
    {\includegraphics[width=0.9\textwidth]{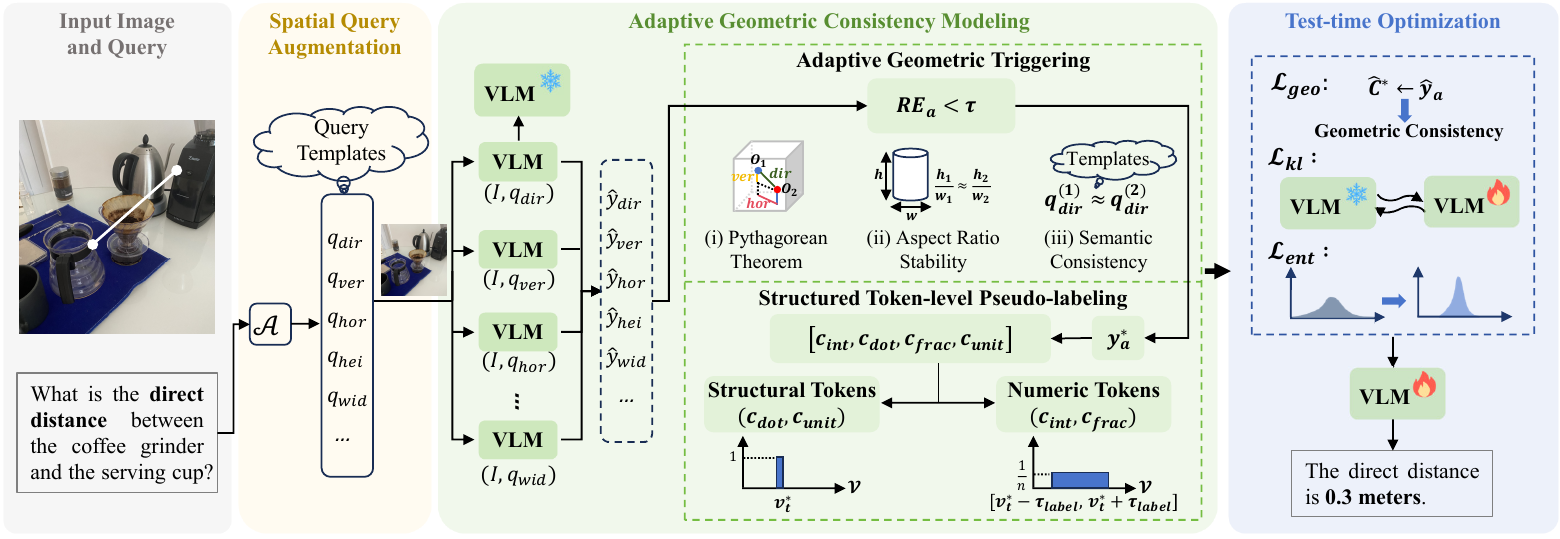}
    }
    \caption{Overview of the TTL-SR pipeline. The framework operates in three phases: (1) Spatial Query Augmentation generates geometrically coupled auxiliary queries via predefined templates; (2) Adaptive Geometric Consistency Modeling filters raw numerical estimates through reliability triggers to construct structured token-level pseudo-labels; and (3) Test-Time Optimization iteratively updates VLM parameters through a multi-objective loss.}
    \label{method_overview}
\end{figure*}

\section{Geometry-aware TTL for Spatial Reasoning}
\subsection{Problem Formulation and Motivation}\label{motivation}
\subsubsection{\textbf{Test-time Learning for Spatial Reasoning}} In quantitative spatial reasoning, a VLM $\Phi_{\theta}(\cdot)$ takes an image $\mathcal{I}$ and a spatial query $q$ to predict a continuous spatial attribute $\hat{y} = \Phi_{\theta}(\mathcal{I}, q)$. Here, $\hat{y} \in \mathbb{R}^+$ is a natural language response containing a numerical measurement (e.g., distance). Given an unlabeled test dataset $\mathcal{D}_{test} = \{(\mathcal{I}_i, q_i)\}_{i=1}^N$ from a target domain, we aim to adapt the model to target distribution via unsupervised optimization:
\begin{equation}
    \min_{\bar{\theta}} \mathcal{L}(\Phi_{\bar{\theta}}; \mathcal{I}, q), \quad (\mathcal{I}, q) \sim \mathcal{D}_{test},
\end{equation}
where $\bar{\theta} \subseteq \theta$ denotes the subset of model parameters updated during test time. The choice of the objective $\mathcal{L}(\cdot)$ is critical, as it must effectively activate the model's spatial knowledge without access to ground-truth labels.

\subsubsection{\textbf{Motivation for Geometry Consistency Optimization}}
\begin{figure}[h]
\centerline
{\includegraphics[width=0.9\columnwidth]{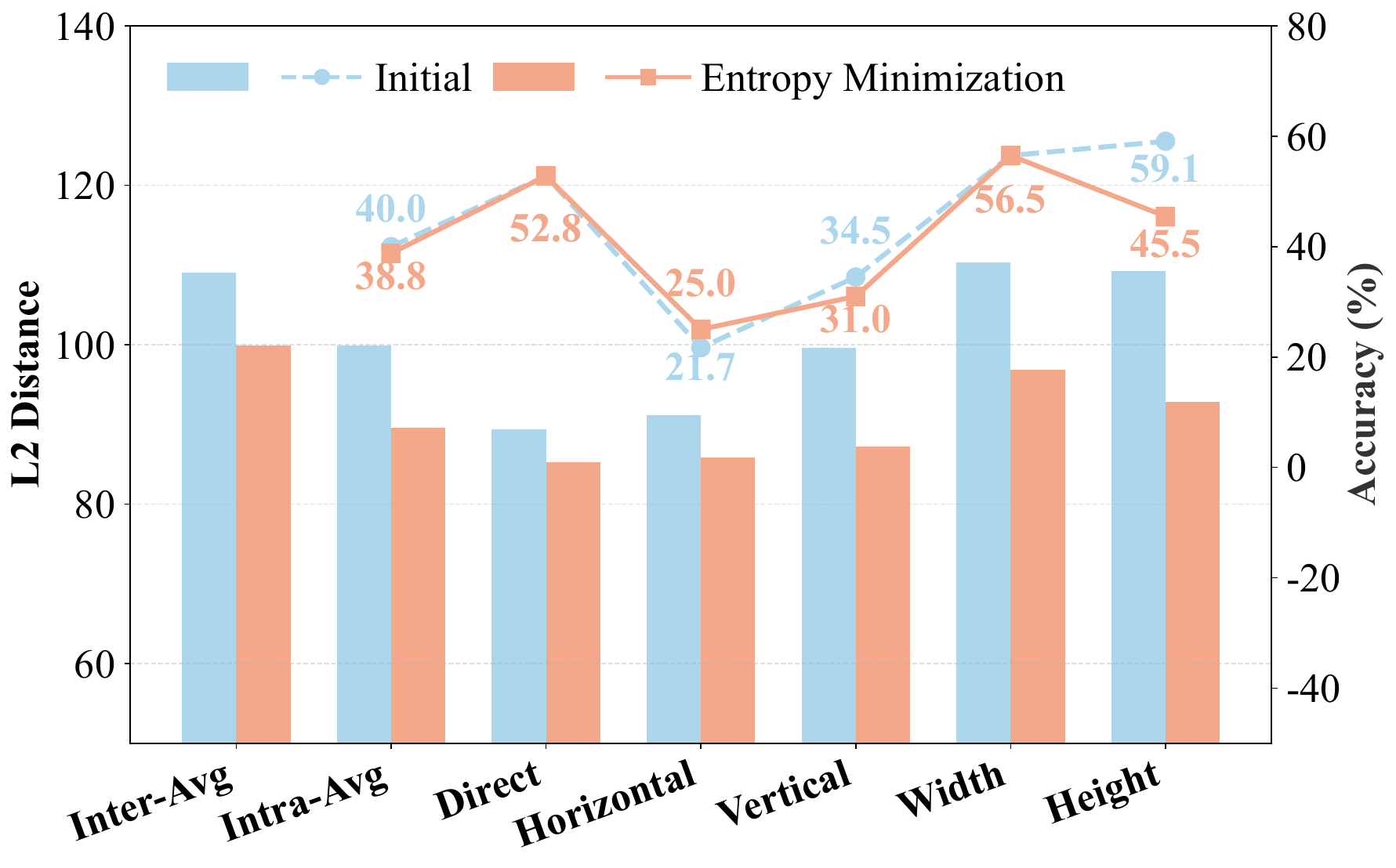}
}
    \caption{Preliminary analysis of feature space evolution and reasoning performance on the Q-Spatial-ScanNet dataset~\cite{liao-etal-2024-reasoning}. Bars (left y-axis) show the average pairwise $L_2$ distance between sample features extracted from the last hidden layer of Qwen3-VL-4B-Instruct~\cite{yang2025qwen3}, while markers and lines (right y-axis) indicate the corresponding accuracy, respectively.}
    \label{distance_acc}
\end{figure}

One straightforward choice for $\mathcal{L}(\cdot)$ is entropy minimization, which has been widely adopted in prior TTL literature~\cite{wang2021tent,niu2022efficient}. In our preliminary studies, we first examined this objective in our quantitative spatial reasoning context. However, we observed that naive entropy minimization yields only marginal improvements, or even degrades performance, as shown in Figure~\ref{distance_acc}.

Upon closer inspection, our analysis reveals also that naively minimizing prediction entropy leads to a rapid increase in feature similarity, indicating a strong mode-seeking behavior in the representation space. Such representation homogenization is beneficial for classification tasks. However, it becomes problematic for quantitative spatial reasoning, where accurate continuous prediction relies on preserving fine-grained geometric relationships between samples. This observation motivates us to move beyond confidence-driven objectives and introduce a geometry-consistent learning objective, which explicitly constrains the relative structure of features during adaptation and prevents representation degeneration. 
Extended analysis are provided in the supplementary materials C.1.

\subsection{Method Overview}
Based on the above motivation, we propose \textbf{TTL-SR}, a geometry-aware test-time learning framework designed to enhance quantitative spatial reasoning in VLMs. The overall pipeline is illustrated in Figure~\ref{method_overview}.
Given an input image $\mathcal{I}$ and a spatial query $q$, the primary challenge is that standard inference treats the query in isolation, leaving the prediction under-constrained and often physically implausible.
To exploit the implicit geometric structure within the scene, we introduce \textbf{Spatial Query Augmentation} (c.f. Sec.~\ref{query_aug}), which expands the original query into a set of geometrically coupled auxiliary queries.
These queries are paired with the same image and processed independently by the VLM to produce raw numerical estimates $\{\hat{y}_{dir}, \hat{y}_{ver}, \hat{y}_{hor}, \dots\}$.

To explicitly leverage geometric consistency as an optimization objective, we introduce \textbf{Adaptive Geometric Consistency Modeling} (Sec.~\ref{adaptive_geometric}). 
For a given set of predictions, we construct a numerical pseudo target $y^*_{a}$ from auxiliary predictions and compute its relative error $RE_{a}$ with respect to the original output $\hat{y}_{a}$, where $a$ denotes the question type. 
The relative error serves as an indicator of internal reasoning coherence: only when the model’s predictions are mutually consistent under geometric constraints (i.e., $RE_{a} < \tau$) do we treat the numerical pseudo target as reliable. In this case, $y^*_{a}$ is used to construct token-level pseudo-labels $\hat{C}^*$, effectively converting the numerical target into a discrete sequence that provides a structured, learnable self-supervision signal.

Finally, in \textbf{Test-Time Optimization} (Sec.~\ref{test_time_optimization}), the model is optimized using these pseudo-labels by minimizing:
\begin{equation}
    \mathcal{L}_{\text{total}} = \mathcal{L}_{\text{geo}} + \lambda_{\text{kl}} \mathcal{L}_{\text{kl}} + \lambda_{\text{ent}} \mathcal{L}_{\text{ent}},
\end{equation}
where $\mathcal{L}_{\text{geo}}$ leverages geometric constraints via the generated pseudo-labels, $\mathcal{L}_{\text{kl}}$ mitigates distributional drift to safeguard against representation collapse, and $\mathcal{L}_{\text{ent}}$ regularizes predictive certainty.
By iteratively minimizing $\mathcal{L}_{\text{total}}$ at test time, TTL-SR encourages physically plausible and self-consistent spatial predictions.

\subsection{Spatial Query Augmentation}\label{query_aug}
Standard VLM inference processes spatial queries independently, ignoring geometric dependencies among attributes and leaving numerical predictions under-constrained.
To expose these dependencies, we introduce a spatial query augmentation operator $\mathcal{A}$ that expands an isolated query into a set of coupled auxiliary queries.
Given a target query $q$, $\mathcal{A}$ generates related variants that probe complementary components of the same scene.

For example, consider the direct distance query: “What is the direct distance between the coffee grinder and the coffee serving cup?”. 
We construct auxiliary queries that decompose this relation into geometric components, such as “What is the vertical distance between the coffee grinder and the coffee serving cup?” and “What is the horizontal distance between the coffee grinder and the coffee serving cup?”, forming $\mathcal{Q}_{rel}=\{q_{dir}, q_{ver}, q_{hor}\}$.
We further add intrinsic attribute queries, e.g., “What is the height of the coffee grinder?” and “What is the width of the coffee serving cup?”, forming the final query set $\mathcal{Q}=\mathcal{Q}_{rel}\cup\mathcal{Q}_{attr}$.
All queries are generated via template-based transformations without external supervision. 
Querying the VLM with $\mathcal{Q}$ induces a set of geometrically coupled predictions, whose internal consistency provides a structured self-supervision signal for subsequent test-time adaptation.

\subsection{Adaptive Geometric Consistency Modeling}\label{adaptive_geometric}
Based on the augmented query set $\mathcal{Q}$, we independently probe the VLM to obtain a set of raw numerical estimates. 
These estimates provide the metric scale through the base VLM, while geometric constraints only relate coupled quantities, such as direct, vertical, and horizontal distances. 
Thus, our goal is not to infer absolute scale from geometry alone, but to convert reliable cross-query agreement into self-supervision.
The primary challenge is determining when such consistency signals are reliable enough for optimization. 
Although geometric principles (e.g., the Pythagorean theorem) provide natural constraints, directly enforcing them can be counterproductive: prediction errors and hallucinations may violate these relations, leading to unstable gradients and error amplification.

To address this issue, we introduce Adaptive Geometric Consistency Modeling, which treats geometric relations as a reliability-aware filtering mechanism. 
Instead of directly enforcing geometric constraints, we use them to assess the internal consistency of predictions and selectively enable optimization. 
This design ensures that adaptation is guided by physically plausible reasoning patterns, while preventing updates driven by erroneous or noisy estimates.

\subsubsection{\textbf{Adaptive Geometric Triggering}}
Before constructing optimization targets, we first check whether the VLM predictions are internally coherent enough to support self-supervision. We therefore introduce a lightweight geometric trigger that activates test-time optimization only for reliable prediction groups.

Specifically, we evaluate a small set of consistency conditions expected in physically plausible scenes: (i) \emph{Pythagorean consistency}, where the predicted direct distance $\hat{y}_{dir}$ should agree with its vertical and horizontal components via $y^*_{dir} = \sqrt{\hat{y}_{ver}^2 + \hat{y}_{hor}^2}$; (ii) \emph{aspect ratio stability}, where relative proportions such as width-to-height ratios remain stable across equivalent queries; and (iii) \emph{semantic consistency}, where predictions for the same spatial attribute under different natural language formulations remain mutually consistent.

Rather than serving as direct supervision, these conditions provide reliability criteria for selecting valid optimization signals. For a target attribute $a$, we derive a reference value $y^*_a$ from its coupled predictions and trigger test-time optimization only when its relative error to the model prediction $\hat{y}_a$ satisfies:
\begin{equation}
    \mathrm{RE}_a = \frac{|\hat{y}_a - y^*_a|}{y^*_a} < \tau,
\end{equation}
where $\tau$ is a predefined tolerance threshold.



\subsubsection{\textbf{Structured Token-level Pseudo-labeling}}\label{pseudo-labeling}
Once the geometric triggering condition is satisfied, we convert the reconciled estimate $y^*_a$ into a token-level optimization target. This step is essential, as geometric constraints are defined in continuous numerical space and cannot be directly optimized in autoregressive VLMs.
Before tokenization, we express $y^*_a$ in the same unit as the model's original prediction $\hat{y}_a$, avoiding mismatched supervision such as centimeters versus meters.
We represent each quantitative answer as a structured token sequence $S = [c_{int}, c_{dot}, c_{frac}, c_{unit}]$,
corresponding to the integer part, decimal point, fractional part, and measurement unit. This decomposition follows the standard format of numerical expressions, when certain components are absent (e.g., no fractional part or unit), the sequence is deterministically adjusted to maintain consistent token alignment.
For each token position $t$, we construct a pseudo-label distribution $\hat{C}^*_t$ over the vocabulary $\mathcal{V}$ based on $y^*_a$.

\paragraph{\textbf{Structural tokens}} For structural tokens ($c_{dot}$, $c_{unit}$), which determine numerical formatting and physical scale, we apply deterministic anchoring using one-hot pseudo-labels:
\begin{equation}
    \hat{C}^*_t(v) =
    \begin{cases}
    1, & v = v^*_t \\
    0, & \text{otherwise},
    \end{cases}
    \quad \forall v \in \mathcal{V},
\end{equation}
where $v^*_t$ denotes the target token at position $t$.
The decimal point token is predefined by the output format, while the unit token is inherited from the model's current prediction $\hat{y}_a$.

\paragraph{\textbf{Numeric tokens}} For numeric tokens ($c_{int}$, $c_{frac}$), we explicitly model uncertainty in $y^*_a$. Let $v^*_t$ denote the digit token corresponding to $y^*_a$ at position $t$.
We define a tolerance interval $[v^*_t - \tau_{label},\, v^*_t + \tau_{label}]$ and collect all tokens within this range. Formally, let $\mathcal{V}_{nbr} \subset \mathcal{V}_{num} \subset \mathcal{V}$ denote this candidate set.
We then construct a uniform pseudo-label distribution:
\begin{equation}
    \hat{C}^*_t(v) =
    \begin{cases}
    \frac{1}{|\mathcal{V}_{nbr}|}, & v \in \mathcal{V}_{nbr} \\
    0, & \text{otherwise},
    \end{cases}
\end{equation}
where $|\mathcal{V}_{nbr}|$ denotes the number of valid tokens.
This soft supervision distributes probability mass over a plausible numerical interval rather than collapsing to a single value, yielding stable gradients that are robust to noise in geometry-derived estimates and enabling reliable test-time learning.

\subsection{Test-Time Optimization}\label{test_time_optimization}

Quantitative spatial answers are expressed as structured numerical sequences.
Given an image-query pair $(\mathcal{I}, q)$, the model produces a logit tensor $\mathbf{Z} \in \mathbb{R}^{L \times |\mathcal{V}|}$, where $L$ is the output sequence length. Following Sec.~\ref{pseudo-labeling}, 
we select geometry-relevant token positions 
$S = \{s_1, s_2, s_3, s_4\}$ corresponding to $[c_{int}, c_{dot}, c_{frac}, c_{unit}]$, and restrict optimization to these positions:
\begin{equation}
    \mathbf{Z}_S = [\mathbf{z}_{s_1}, \mathbf{z}_{s_2}, \mathbf{z}_{s_3}, \mathbf{z}_{s_4}]^\top,
\end{equation}
where $\mathbf{z}_{s_i} \in \mathbb{R}^{|\mathcal{V}|}$ denotes the logits at position $s_i$.
For each $s_i \in S$, the predictive distribution is obtained via softmax:
\begin{equation}
    P_{\theta}(v \mid \mathcal{I}, q, t = s_i)
    = \frac{\exp(\mathbf{z}_{s_i}^{(v)})}
    {\sum_{v' \in \mathcal{V}} \exp(\mathbf{z}_{s_i}^{(v')})},
\end{equation}
this design ensures that gradients are propagated only through tokens that directly determine the numerical value and physical scale of spatial predictions.

\subsubsection{\textbf{Geometric Consistency Loss}}
This loss serves as the core supervision signal and is composed of complementary components. We adopt MLE Positive Learning and employ a multi-label cross-entropy loss \cite{zhang2018generalized} to encourage the model to distribute probability mass over all geometrically valid candidate tokens $\mathcal{V}_{nbr}$ at each key position $s_i \in S$. Specifically, for each key position $s_i$, we define the local MLE loss as:
    \begin{equation}
        \ell_{\text{mle}}(s_i)
        = - \sum_{v \in \mathcal{V}_{nbr}}
        \hat{C}^*_{s_i}(v)\,
        \log P_{\theta}(v \mid \mathcal{I}, q, s_i),
    \end{equation}
    where $\hat{C}^*_{s_i}(v)$ denotes the pseudo-target distribution over valid candidate tokens,
    and $P_{\theta}(v \mid \mathcal{I}, q, s_i)$ is the model’s full-vocabulary prediction at position $s_i$.


To suppress confident but geometrically invalid predictions, we apply an Unlikelihood Loss (UL) on numeric positions $S_{ul} \subset S$. 
For each $s_i \in S_{ul}$, a negative token $v_{neg} \notin \mathcal{V}_{nbr}$ is sampled from top-$k$:
\begin{equation}
    \ell_{\text{ul}}(s_i)
    = - (1 - P_{s_i}^{(v_{neg})}) \log (1 - P_{s_i}^{(v_{neg})}),
\end{equation}
the overall geometric consistency loss is:
\begin{equation}
    \mathcal{L}_{\text{geo}}
    = \frac{1}{W} \sum_{s_i \in S} w_i
    \left[
    \ell_{\text{mle}}(s_i)
    + \mathbb{I}(s_i \in S_{ul}) \, \alpha_i \, \ell_{\text{ul}}(s_i)
    \right],
\end{equation}
where $W = \sum_{s_i \in S} w_i$, $w_i$ denotes the position-specific weight that assigns different levels of importance to the measurement components. $\mathbb{I}(\cdot)$ restricts UL to numeric positions, and $\alpha_i = 1/\sqrt{|\mathcal{V}_{nbr}|}$.

\subsubsection{\textbf{KL Loss.}}
To prevent excessive deviation from the pre-trained model, we introduce a global KL-divergence loss \cite{hinton2015distilling}:
\begin{equation}
    \mathcal{L}_{\text{kl}}
    = \frac{1}{L} \sum_{t=1}^{L}
    D_{\text{KL}} \left(
    \sigma(\mathbf{Z}_0^t) \parallel \sigma(\mathbf{Z}^t)
    \right),
\end{equation}
where $\sigma(\cdot)$ denotes the softmax function, $L$ is the sequence length, and $\mathbf{Z}_{0}^t$ and $\mathbf{Z}^t$ represent the logit vectors at position $t$ from the frozen pre-trained model and the current updated model, respectively. This anchors the adapted model to its original output distribution over the full sequence.

\subsubsection{\textbf{Entropy Minimization Loss}}
As discussed in Section~\ref{motivation}, entropy minimization alone may lead to confirmation bias and over-confident but geometrically incorrect predictions. We adopt entropy minimization in a restricted, auxiliary manner. Specifically, it is applied only to the geometry-relevant positions $S = \{s_i\}_{i=1}^{4}$ and is jointly optimized with the geometric consistency and KL loss.
Guided by the pseudo-label distribution $\hat{C}^*$ and the constrained candidate set $\mathcal{V}_{nbr}$, entropy minimization serves to sharpen the model's belief within an already validated geometric interval. 
Formally, the entropy minimization loss is defined as:
\begin{equation}
    \mathcal{L}_{\text{ent}}
    = \frac{1}{|S|} \sum_{s_i \in S}
    \left(
    - \sum_{v \in \mathcal{V}}
    P_{s_i}^{(v)} \log P_{s_i}^{(v)}
    \right),
\end{equation}
$P_{s_i}^{(v)}$ denotes the predicted probability of token $v$ at position $s_i$.

\subsubsection{\textbf{Parameter Optimization}}
Following Sec.~\ref{motivation}, we restrict trainable parameters $\bar{\theta} \subseteq \theta$ to low-rank adapter weights $\Delta \Phi$ injected into the backbone via Low-Rank Adaptation (LoRA) \cite{hu2022lora}.
This design enables efficient test-time adaptation while mitigating catastrophic forgetting, which is commonly observed in full-parameter updates during TTL~\cite{pmlr-v267-hu25z}.
During TTL, the backbone parameters $\theta$ remain frozen and only the adapter weights $\Delta \Phi$ are updated. The optimization proceeds iteratively as:
\begin{equation}
\Delta \Phi_{k+1}
= \Delta \Phi_{k}
- \eta \nabla_{\Delta \Phi}
\mathcal{L}(\Phi_{\theta, \Delta \Phi}; \mathcal{I}, q),
\end{equation}
where $k$ is the optimization step, $\eta$ is the learning rate, and $\Phi_{\theta, \Delta \Phi}$ denotes the frozen backbone with trainable adapters.
Adapters are updated cumulatively over the unlabeled test stream rather than reset per sample; each valid instance contributes one step, and later predictions use the adapted adapters.

\begin{table*}[t]
\centering
\small
\renewcommand{\arraystretch}{0.8}
\caption{Quantitative comparison on Q-Spatial-ScanNet, Q-Spatial++, and the SPAR-Bench object-to-object distance task. Results are reported as Accuracy (\%) $\uparrow$ / MARE $\downarrow$. The Average column denotes the mean accuracy across all categories.}
\label{tab:main_results_Qspatial}
\begin{tabular}{l|cccccc|c|c}
\toprule
\multirow{2}{*}{\textbf{Method}} & \multicolumn{6}{c|}{\textbf{Q-Spatial-ScanNet}} & \multicolumn{1}{c|}{\textbf{Q-Spatial++}}& \multicolumn{1}{c}{\textbf{SPAR-Bench}} \\
\cmidrule(lr){2-7}\cmidrule(lr){8-8}\cmidrule(lr){9-9}
& \textbf{Direct Dist.} & \textbf{Horiz. Dist.} & \textbf{Vert. Dist.} & \textbf{Width} & \textbf{Height} & \textbf{Average} & \textbf{Average} & \textbf{Dist. oo Average} \\ 
\midrule
Doubao-Seed-1.8 
& 22.2 / 0.84 & 6.7 / 3.44 & 13.8 / 1.40 & 56.5 / 0.43 & 59.1 / 0.24 & 24.71 & 22.77 & 44.6 / 0.37 \\
GPT-5.2 
& 47.2 / 0.38 & \textbf{35.0} / 1.12 & 34.5 / 0.70 & \textbf{60.9 / 0.29} & 63.6 / 0.29 & 44.71 & 26.73 &49.4 / 0.29  \\
Gemini-3-Flash 
& \textbf{58.3 / 0.24} & 28.3 / \textbf{0.58} & \textbf{44.8 / 0.66} & 60.9 / 0.34 & \textbf{90.9 / 0.13} & \textbf{50.00} & \textbf{38.61} & \textbf{58.4 / 0.28}\\
Qwen3-VL-Plus 
& 38.9 / 0.52 & 16.7 / 1.74 & 24.1 / 1.86 & 56.5 / 0.36 & 77.3 / 0.21 & 35.88 & 23.76 & 42.1 / 0.41\\
Qwen3-VL-2B-Instruct 
& 38.9 / 0.31 & 31.7 / 1.30 & 20.7 / 0.66 & 52.2 / 0.38 & 40.9 / 0.55 & 35.29 & 31.68 & 39.9 / 0.37 \\
\midrule
Qwen3-VL-4B-Instruct 
& 52.8 / 0.34 & 21.7 / 1.94 & 34.5 / 1.33 & 56.5 / 0.44 & 59.1 / 0.46 & 40.00 & 24.75 & 50.7 / 0.33 \\
+Tent 
&52.8 / 0.30 & 21.7 / 1.43 & 37.9 / 1.43 & 56.5 / 0.49 & 40.9 / 0.44 & 38.24 &30.69 & 52.1 / 0.31\\
+COME 
&52.8 / 0.29 & 25.0 / \textbf{1.28} & \textbf{41.4} / 1.37 & 56.5 / 0.49 & 40.9 / 0.46 & 40.00 &29.70 & 52.3 / 0.31\\
+TLM 
&55.6 / 0.29 & 25.0 / 1.68 & 31.0 / 1.32 & 60.9 / 0.44 & 40.9 / 0.51 & 39.41 & 27.72 & 53.2 / 0.31 \\
\textbf{+TTL-SR (Ours)} 
& \textbf{63.9 / 0.24} & \textbf{28.3} / 1.48 & 34.5 / \textbf{1.23} & \textbf{60.9 / 0.43} & \textbf{68.2 / 0.31} & \textbf{46.47}  & \textbf{34.65} &\textbf{55.1 / 0.30}  \\
\midrule
SpatialRGPT-VILA-1.5-8B 
&47.2 / 0.27 & 26.7 / 0.99 & 24.1 / 1.25 & \textbf{52.2 / 0.41} & 27.3 / 0.58 & 34.12 & 29.70 & -\\
+Tent 
&44.4 / 0.24 & 30.0 / 1.12 & 34.5 / 1.22 & 52.2 / 0.42 & \textbf{54.6} / 0.44 & 40.00  &27.72 & -\\
+COME 
&52.8 / 0.25 & 30.0 / 1.29 & 34.5 / 1.26 & 52.2 / 0.44 & \textbf{54.6} / 0.45 & 41.76 &26.73 & -\\
+TLM 
&52.8 / 0.23 & 33.3 / \textbf{0.89} & 31.0 / 1.22 & 47.8 / 0.43 & 45.5 / 0.56 & 40.60 &30.69 & -\\
\textbf{+TTL-SR (Ours)}  
&\textbf{58.3 / 0.23} & \textbf{35.0} / 1.07 & \textbf{34.5 / 1.18} & 47.8 / 0.45 & 50.0 / \textbf{0.42} & \textbf{43.53} & \textbf{33.66} & -\\
\bottomrule
\end{tabular}
\end{table*}

\section{Experiment}
\subsection{Experimental Settings}
\paragraph{\textbf{Datasets.}} We evaluate TTL-SR on benchmarks covering diverse quantitative spatial reasoning tasks, including distance, object size, and orientation. We primarily use three datasets. 
Firstly, SpatialRGPT-Bench~\cite{cheng2024spatialrgpt} is a spatial VQA benchmark spanning indoor, outdoor, and simulated scenes, with region annotations and quantitative questions for metric estimation.
Secondly, Q-Spatial Bench~\cite{liao-etal-2024-reasoning} includes two subsets: Q-Spatial-ScanNet, based on precise RGB-D measurements from ScanNet~\cite{dai2017scannet}, and Q-Spatial++, containing diverse real-world images under varied conditions. For compatibility with region-aware models, we annotate target object regions for Q-Spatial Bench following the format of SpatialRGPT-Bench.
Thirdly, we incorporate the object-to-object distance prediction task from SPAR-Bench~\cite{zhang2025from}, which focuses on fine-grained spatial relationship estimation between object pairs and serves as a complementary evaluation for metric-level reasoning.
Additional dataset details are provided in the supplementary materials B.1.

\paragraph{\textbf{Evaluation metrics.}} Following~\cite{cheng2024spatialrgpt}, we adopt two primary metrics after normalizing all predictions to a consistent unit:
(i) Accuracy: a prediction $\hat{y}$ is considered correct if its relative error w.r.t.\ the ground truth $y_{gt}$ satisfies $\frac{|\hat{y} - y_{gt}|}{y_{gt}} \le 0.25$. For clock-face direction, we use an absolute tolerance of 2 hours.
(ii) Mean Absolute Relative Error (MARE): we report the average absolute relative error over all samples to measure overall measurement precision.

\paragraph{\textbf{VLMs and baselines.}} We evaluate TTL-SR on two representative open-source VLM backbones: Qwen3-VL-4B-Instruct~\cite{yang2025qwen3} and SpatialRGPT-VILA-1.5-8B~\cite{cheng2024spatialrgpt}.
We compare TTL-SR against representative TTA or TTL methods, including Tent~\cite{wang2021tent}, COME~\cite{zhang2024come}, and TLM~\cite{pmlr-v267-hu25z}, which adapt model parameters using unlabeled data at inference time.
In addition, we also compare with frontier closed-source models, including Doubao-Seed-1.8~\cite{seed1.8}, GPT-5.2~\cite{openai2025gpt52}, Gemini-3-Flash~\cite{google2025gemini3}, Qwen3-VL-Plus~\cite{yang2025qwen3}. Additional details are provided in the supplementary materials B.1.

\begin{table*}[t]
\centering
\small
\renewcommand{\arraystretch}{0.8}
\caption{Quantitative comparison of spatial reasoning performance on SpatialRGPT-Bench. Results are reported as Accuracy (\%) $\uparrow$ / MARE $\downarrow$. The Direction metric are reported as Accuracy (\%)$\uparrow$ / Mean Angular Error ($^\circ$)$\downarrow$.}
\label{tab:main_results_spatialbench}
\begin{tabular}{l|ccccccc}
\toprule
\textbf{Method} & \textbf{Direct Dist.} & \textbf{Horiz. Dist.} & \textbf{Vert. Dist.} & \textbf{Width} & \textbf{Height} & \textbf{Direction} & \textbf{Average}  \\ 
\midrule
Doubao-Seed-1.8 & 24.3 / 0.67 & 21.3 / 1.38 & \textbf{33.0} / 0.52 & 57.1 / 0.32 & 68.4 / 2.90 & 55.1 / 55.79$^{\circ}$ & 43.12 \\
GPT-5.2 & 15.5 / 0.57 & 27.1 / 1.14 & 20.8 / \textbf{0.46} & 59.7 / 0.33 & \textbf{71.4} / 1.91 & \textbf{64.5 / 42.06$^{\circ}$} & 42.72 \\
Gemini-3-Flash & 17.6 / 0.59 & 31.2 / 0.75 & 17.9 / 0.52 & \textbf{63.9 / 0.29} & 68.4 / \textbf{1.77} & 57.0 / 47.10$^{\circ}$ & 42.72 \\
Qwen3-VL-Plus & \textbf{31.8 / 0.48} & \textbf{32.0} / 1.11 & 31.1 / 0.56 & 51.1 / 0.38 & 69.2 / 1.79 & 49.1 / 60.00$^{\circ}$ & \textbf{44.25} \\
Qwen3-VL-2B-Instruct & 16.2 / 0.55 & 30.3 / \textbf{0.51} & 17.0 / 0.61 & 32.3 / 0.45 & 37.6 / 2.10 & 30.7 / 78.10$^{\circ}$ & 27.32 \\
\midrule
Qwen3-VL-4B-Instruct & 29.1 / 0.51 & 35.3 / 0.96 & 29.3 / 0.74 & 39.9 / 0.42 & 64.7 / 1.25 & 41.1 / 61.70$^{\circ}$ & 40.05 \\
+Tent &33.1 / 0.49 & 40.2 / 0.92 & 31.1 / 0.65 & 36.8 / 0.40 & 63.2 / 1.23 & 43.9 / 61.12$^{\circ}$ & 41.52 \\
+COME &30.4 / 0.49 & 42.6 / 0.91 & 32.1 / 0.66 & 38.4 / 0.41 & 66.2 / 1.23 & 39.3 / 63.08$^{\circ}$ & 41.66 \\
+TLM &34.5 / 0.49 & 42.6 / 0.94 & \textbf{33.0} / 0.65 & 38.4 / 0.42 & 66.9 / 1.32 & 38.3 / 64.77$^{\circ}$ & 42.59 \\
\textbf{+TTL-SR (Ours)} & \textbf{35.8 / 0.47} & \textbf{42.6 / 0.87} & 31.1 / \textbf{0.63} & \textbf{41.4 / 0.40} & \textbf{66.9 / 1.21} & \textbf{45.8 / 60.84$^{\circ}$} & \textbf{44.19} \\
\midrule
SpatialRGPT-VILA-1.5-8B & 45.9 / \textbf{0.31} & \textbf{68.0 / 0.22} & 56.6 / 0.28 & 48.9 / \textbf{0.28} & 61.7 / 0.41 & 95.3 / \textbf{9.70$^{\circ}$} & - \\
+Tent &38.5 / 0.35 & 62.3 / 0.23 & 60.4 / \textbf{0.27} & 51.1 / 0.31 & \textbf{72.9} / 0.40 & \textbf{96.3} / 10.37$^{\circ}$ & 62.08 \\
+COME &45.3 / 0.34 & 59.8 / 0.23 & 55.7 / 0.29 & 52.6 / 0.33 & 72.2 / 0.39 & \textbf{96.3} / 11.21$^{\circ}$ & 62.48 \\
+TLM &45.3 / 0.34 & 57.4 / 0.25 & 53.8 / 0.29 & 51.9 / 0.33 & 66.9 / 0.38 & \textbf{96.3} / 11.78$^{\circ}$ & 60.75 \\
\textbf{+TTL-SR (Ours)} &\textbf{46.6} / 0.32 & 60.7 / 0.23 & \textbf{64.2} / 0.28 & \textbf{54.9} / 0.33 & 70.7 / \textbf{0.37} & \textbf{96.3} / 11.20$^{\circ}$ & \textbf{64.22} \\
\bottomrule
\end{tabular}
\end{table*}

\paragraph{\textbf{Implementation details.}} We evaluate TTL-SR on Qwen3-VL-4B-Instruct~\cite{yang2025qwen3} and SpatialRGPT-VILA-1.5-8B~\cite{cheng2024spatialrgpt}, leveraging Qwen-VL finetuning framework~\cite{Qwen-VL-Finetuning} and official SpatialRGPT~\cite{cheng2024spatialrgpt} codebases.
Unless stated otherwise, we set $\lambda_{kl}=50$, $\lambda_{ent}=0.2$, and $\tau=\tau_{label}=0.20$.
Negative tokens for UL loss are selected via top-$k$ search with $k=10$.
Greedy decoding is used throughout. Experiments are conducted
on NVIDIA RTX 4090 (24GB) for the 4B model and NVIDIA A100
(80GB) for the 8B model.
Additional details are provided in supplementary materials B.2.

\subsection{Comparison Experiments}\label{main_experiments}
We compare TTL-SR with SOTA closed-source VLMs, original open-source counterparts, and representative TTA / TTL methods (Tent, COME, TLM). Experiments are conducted on Q-Spatial-ScanNet, Q-Spatial++, the object-to-object distance subset of SPAR-Bench, and SpatialRGPT-Bench, as shown in Table~\ref{tab:main_results_Qspatial} and Table~\ref{tab:main_results_spatialbench}.

\subsubsection{\textbf{Comparison with Original VLMs}} TTL-SR consistently outperforms original backbones across all benchmarks. On Q-Spatial-ScanNet, it yields a 6.47\% absolute accuracy gain for Qwen3-VL-4B-Instruct, with consistent MARE reductions across all query categories. Notably, TTL-SR demonstrates strong generalization for specialized region-aware models. For instance, it improves SpatialRGPT-VILA-1.5-8B by 9.41\% in overall accuracy. Similar improvements are observed on the SPAR-Bench distance subset, where TTL-SR enhances fine-grained object-to-object distance estimation. These results verify that TTL-SR provides a robust and architecture-agnostic improvement for quantitative spatial reasoning.

\subsubsection{\textbf{Comparison with TTA / TTL Methods}}
As shown in Table~\ref{tab:main_results_Qspatial} and Table~\ref{tab:main_results_spatialbench}, existing TTA / TTL methods exhibit unstable performance in quantitative spatial reasoning. When adapting Qwen3-VL-4B-Instruct on Q-Spatial-ScanNet, Tent and TLM actually degrade the model: average accuracy drops from 40.00\% to 38.24\% and 39.41\%, respectively, while Tent increases vertical distance MARE from 1.33 to 1.43.
In contrast, TTL-SR consistently yields substantial gains across all metrics. It reduces vertical distance MARE to 1.23 and boosts the average accuracy to 46.47\%, demonstrating both higher precision and better stability.
Interestingly, on the SpatialRGPT-VILA-1.5-8B backbone evaluated on SpatialRGPT-Bench, all methods see a regression in horizontal distance. One possible reason is that the model's strong pre-trained priors for this specific dimension are particularly sensitive to representation drift during test-time optimization.
Overall, these results suggest that confidence-driven adaptation is insufficient for reliable numeric refinement, whereas explicitly introducing geometric consistency enables more stable and effective test-time optimization.

\subsubsection{\textbf{Comparison with Closed-source VLMs}}
As shown in Table~\ref{tab:main_results_Qspatial} and \ref{tab:main_results_spatialbench}, quantitative spatial reasoning remains a challenging problem even for frontier closed-source models such as GPT-5.2 and Gemini-3-Flash. While these massive models excel in width and height estimation by leveraging extensive object-level priors, they struggle with complex distance queries. In contrast, our lightweight Qwen3-VL-4B-Instruct with TTL-SR achieves competitive or superior performance across multiple benchmarks. For example, on Q-Spatial-ScanNet, it reaches 63.9\% accuracy on direct distance, surpassing the 58.3\% achieved by Gemini-3-Flash. These results demonstrate that incorporating explicit geometric consistency at test time can effectively bridge the performance gap between compact open-source models and large-scale closed-source VLMs.

\begin{figure*}[t]
\centerline{
\includegraphics[width=0.9\textwidth]{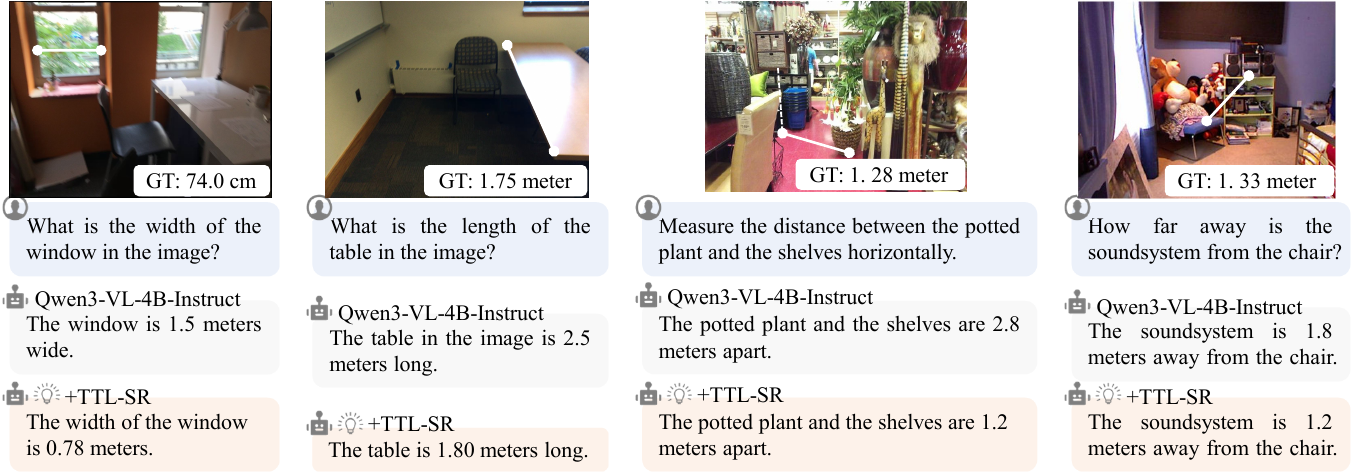}
}
    \caption{Qualitative comparison of spatial reasoning between the baseline Qwen3-VL-4B-Instruct and our TTL-SR.}
    \label{examples}
\end{figure*}

\subsection{Ablation Study}
To evaluate the contribution of individual components in TTL-SR, we conduct a series of ablation experiments on the Q-Spatial-ScanNet dataset using the Qwen3-VL-4B-Instruct backbone. The results are summarized in Table~\ref{tab:ablation_qwen_loss}, Table~\ref{tab:ablation_qwen_geo}, Table~\ref{tab:ablation_geo_terms}, and Table~\ref{tab:ablation_paramters}.

\begin{table}[h]
\centering
\renewcommand{\arraystretch}{0.95}
\caption{Ablation study of TTL-SR loss components.}
\label{tab:ablation_qwen_loss}
\resizebox{\linewidth}{!}{
\begin{tabular}{ccc|cccccc}
\toprule
\textbf{Ent} & \textbf{KL} & \textbf{Geo} & \makecell[c]{\textbf{Direct}\\\textbf{Dist.}} & \makecell[c]{\textbf{Horiz.}\\\textbf{Dist.}} & \makecell[c]{\textbf{Vert.}\\\textbf{Dist.}} & \textbf{Width} & \textbf{Height} & \textbf{Average} \\
\midrule
 &  &  & 52.8 / 0.34 & 21.7 / 1.94 & 34.5 / 1.33 & 56.5 / 0.44 & 59.1 / 0.46 & 40.00 \\
\checkmark &  &  & 58.3 / 0.28 & 23.3 / \textbf{1.44} & \textbf{37.9 / 1.20} & 60.9 / 0.50 & 45.5 / 0.44 & 40.18 \\
\checkmark & \checkmark &  & 55.6 / 0.29 & 21.7 / 1.61 & 37.9 / 1.23 & 60.9 / 0.49 & 50.0 / 0.44 & 40.59 \\
 & \checkmark & \checkmark & 55.6 / 0.25 & \textbf{28.3} / 1.62 & 37.9 / 1.38 & 52.2 / 0.46 & 59.1 / 0.41 & 42.94 \\
\checkmark &  & \checkmark & 61.1 / 0.25 & 25.0 / 1.58 & 37.9 / 1.25 & 52.2 / 0.45 & 63.6 / \textbf{0.30} & 43.53 \\
\checkmark & \checkmark & \checkmark & \textbf{63.9 / 0.24} & \textbf{28.3} / 1.48 & 34.5 / 1.23 & \textbf{60.9 / 0.43} & \textbf{68.2} / 0.31 & \textbf{46.47} \\
\bottomrule
\end{tabular}}
\end{table}

\subsubsection{\textbf{Effectiveness of Loss Components}}
We first investigate the impact of different loss functions on model performance. As shown in Table~\ref{tab:ablation_qwen_loss}, applying entropy minimization or KL regularization in isolation yields only marginal improvements over the baseline. 
Specifically, while entropy minimization slightly increases the average accuracy from 40.00\% to 40.18\% , it introduces instability across certain spatial categories, such as a performance degradation in height estimation. Incorporating KL regularization enhances optimization stability and raises the average accuracy to 40.59\%. However, the overall gains remain constrained, suggesting that confidence-based objectives alone are insufficient for mastering complex quantitative spatial reasoning.

In contrast, the integration of our proposed geometric consistency loss leads to substantial performance gains across all spatial categories. When combined with KL regularization, the average accuracy increases to 42.94\%, and further reaches 46.47\% when all components are jointly optimized, representing a 6.47\% absolute improvement over the baseline. While entropy and KL losses help stabilize test-time optimization, the primary performance boost comes from explicit geometric consistency. This confirms that geometric priors are indispensable for the model to reason accurately about quantitative spatial reasoning.

\begin{table}[h]
\centering
\small
\renewcommand{\arraystretch}{0.95}
\caption{Impact of individual geometric constraints.}
\label{tab:ablation_geo_terms}
\resizebox{\linewidth}{!}{
\begin{tabular}{l|cccccc}
\toprule
\textbf{Constraints} & \makecell[c]{\textbf{Direct}\\\textbf{Dist.}} & \makecell[c]{\textbf{Horiz.}\\\textbf{Dist.}} & \makecell[c]{\textbf{Vert.}\\\textbf{Dist.}} & \textbf{Width} & \textbf{Height} & \textbf{Average} \\
\midrule
Baseline (Ent+KL) & 55.6 / 0.29 & 21.7 / 1.61 & \textbf{37.9 / 1.23} & 60.9 / 0.49 & 50.0 / 0.44 & 40.59 \\
w/ $\mathcal{L}_{geo}$ (Same)   & 58.3 / 0.26 & 28.3 / 1.61 & 34.5 / 1.42 & 56.5 / 0.46 & 54.6 / 0.41 & 42.94 \\
w/ $\mathcal{L}_{geo}$ (Direct)    & 58.3 / 0.29 & 23.3 / 1.61 & 31.0 / 1.57 & 60.9 / 0.46 & 59.1 / 0.40 & 41.76 \\
w/ $\mathcal{L}_{geo}$ (Vert.)  & 61.1 / 0.27 & 25.0 / 1.66 & 37.9 / 1.30 & 60.9 / 0.47 & 50.0 / 0.44 & 42.94 \\
w/ $\mathcal{L}_{geo}$ (Horiz.)& 58.3 / 0.29 & 26.7 / 1.65 & \textbf{37.9 / 1.23} & 60.9 / 0.50 & 50.0 / 0.44 & 42.94 \\
w/ $\mathcal{L}_{geo}$ (Width)     & 58.3 / 0.28 & 25.0 / 1.64 & 31.0 / 1.39 & 60.9 / 0.48 & 50.0 / 0.44 & 41.18 \\
w/ $\mathcal{L}_{geo}$ (Height)    & 55.6 / 0.29 & 25.0 / 1.54 & 31.0 / 1.45 & \textbf{65.2} / 0.47 & 54.6 / 0.41 & 41.76 \\
w/ $\mathcal{L}_{geo}$(All) & \textbf{63.9 / 0.24} & \textbf{28.3 / 1.48} & 34.5 / 1.23 & 60.9 / \textbf{0.43} & \textbf{68.2 / 0.31} & \textbf{46.47} \\
\bottomrule
\end{tabular}}
\end{table}



\subsubsection{\textbf{Analysis of Geometric Constraints}}
We further decompose the geometric consistency constraints to understand how individual constraints guide spatial reasoning, as depicted in Table~\ref{tab:ablation_geo_terms}. The "Same" constraint ensures consistent outputs across diverse linguistic phrasings, while dimension-specific constraints (direct, height, etc.) anchor the model to fundamental physical principles.

Our results reveal a tight coupling between spatial attributes: optimizing one dimension in isolation may distorts others. For example, activating only the w/ $\mathcal{L}_{geo}$ (Direct) improves direct distance accuracy from 55.6\% to 58.3\%, but simultaneously degrades vertical distance performance from 37.9\% to 31.0\%. TTL-SR bypasses these trade-offs by regularizing all spatial attributes in a unified optimization space. By respecting the entire geometric structure, our framework avoids skewed updates that favor specific dimensions. This highlights the necessity of coordinated geometric priors for achieving stable and physically plausible spatial reasoning.

\begin{table}[h]
\centering
\renewcommand{\arraystretch}{0.95}
\caption{Component-wise ablation of TTL-SR framework.}
\label{tab:ablation_qwen_geo}
\resizebox{\linewidth}{!}{
\begin{tabular}{l|cccccc}
\toprule
\textbf{Method} & \makecell[c]{\textbf{Direct}\\\textbf{Dist.}} & \makecell[c]{\textbf{Horiz.}\\\textbf{Dist.}} & \makecell[c]{\textbf{Vert.}\\\textbf{Dist.}} & \textbf{Width} & \textbf{Height} & \textbf{Average} \\
\midrule
w/o AT &61.1 / 0.25 & 23.3 / 1.50 & 34.5 / 1.27 & 56.5 / 0.44 & 68.2 / 0.32 & 43.53 \\
w/o SPL &58.3 / 0.27 & 28.3 / 1.64 & 27.6 / 1.45 & 60.9 / 0.45 & 63.6 / 0.32 & 43.53 \\
w/o UL &58.3 / 0.25 & 28.3 / 1.57 & 31.0 / 1.46 & \textbf{60.9 / 0.43} & 63.6 / 0.32 & 44.12 \\
Ours  & \textbf{63.9 / 0.24} & \textbf{28.3 / 1.48} & \textbf{34.5 / 1.23} & \textbf{60.9 / 0.43} & \textbf{68.2 / 0.31} & \textbf{46.47} \\
\bottomrule
\end{tabular}}
\end{table}

\begin{table}[h]
\renewcommand{\arraystretch}{0.95}
\caption{Sensitivity analysis of key hyper-parameters in TTL-SR, including the trigger threshold $\tau$, the soft label tolerance interval $\tau_{label}$, and the sampling top-$k$ for UL.}
\label{tab:ablation_paramters}
\resizebox{\linewidth}{!}{
\begin{tabular}{c|c|cccccc}
\toprule
\textbf{Param.} & \textbf{Value} &\makecell[c]{\textbf{Direct}\\\textbf{Dist.}} & \makecell[c]{\textbf{Horiz.}\\\textbf{Dist.}} & \makecell[c]{\textbf{Vert.}\\\textbf{Dist.}} & \textbf{Width} & \textbf{Height} & \textbf{Average} \\
\midrule
\multirow{4}{*}{\makecell[c]{\textbf{$\tau$}}} & 0.00 & 61.1 / 0.25 & 23.3 / 1.50 & 34.5 / 1.27 & 56.5 / 0.44 & 68.2 / 0.32 & 43.53 \\
& 0.10 & 58.3 / 0.26 & 26.7 / 1.52 & \textbf{37.9 / 1.23} & 56.5 / 0.44 & 63.6 / 0.31 & 44.12 \\
& \textbf{0.20} & \textbf{63.9 / 0.24} & \textbf{28.3 / 1.48} & 34.5 / 1.23 & \textbf{60.9 / 0.43} & \textbf{68.2 / 0.31} & \textbf{46.47} \\
& 0.40 & 61.1 / 0.26 & 25.0 / 1.59 & 34.5 / 1.40 & 52.2 / 0.46 & 63.6 / 0.32 & 42.94 \\
\midrule
\multirow{4}{*}{\makecell[c]{\textbf{$\tau_{label}$}}} & 0.10 & 61.1/ 0.25 & 25.0   / 1.56 & 37.9 / 1.25 & 47.8 / 0.46 & 68.2 / 0.33 & 43.53 \\
 & \textbf{0.20} & \textbf{63.9 / 0.24} & \textbf{28.3 / 1.48} & 34.5 / 1.23 & \textbf{60.9 / 0.43} & \textbf{68.2 / 0.31} & \textbf{46.47} \\
& 0.30 & 58.3 / 0.25 & 25.0 / 1.54 & \textbf{41.4 / 1.18} & 56.5 / 0.44 & 63.6 / 0.31 & 44.12 \\
& 0.40 & 61.1 / 0.26 & 23.3 / 1.58 & 37.9 / 1.22 & 56.5 / 0.46 & 63.6 / 0.33 & 43.53 \\
\midrule
\multirow{4}{*}{\textbf{$k$}} & 1 & 58.3 / 0.25 & 26.7 / 1.53 & 37.9 / 1.19 & 52.2 / 0.45 & \textbf{68.2 / 0.31} & 44.12 \\
& 5 & 61.1 / 0.26 & 25.0 / 1.59 & 34.5 / 1.30 & \textbf{60.9 / 0.43} & 63.6 / 0.31 & 44.12 \\
& \textbf{10} & \textbf{63.9 / 0.24} & \textbf{28.3 / 1.48} & 34.5 / 1.23 & \textbf{60.9 / 0.43} & \textbf{68.2 / 0.31} & \textbf{46.47} \\
& 20 & \textbf{63.9 / 0.24} & 26.7 / 1.50 & \textbf{41.4 / 1.15} & 52.2 / 0.45 & 68.2 / 0.32 & 45.88 \\
\bottomrule
\end{tabular}}
\end{table}



\subsubsection{\textbf{Effectiveness of Framework Components}}
Table~\ref{tab:ablation_qwen_geo} breaks down the contribution of each TTL-SR module. Removing any single component leads to a consistent drop in accuracy and a rise in MARE, confirming their collective necessity.
Specifically, omitting Adaptive Triggering (AT) incurs a 2.94\% decline in accuracy, as the model suffers from indiscriminate updates on physically implausible predictions. Replacing Soft Pseudo-Labeling (SPL) with hard one-hot encoding similarly degrades precision. Furthermore, excluding the Unlikelihood Loss (UL) significantly elevates the MARE (e.g., from 1.23 to 1.46 in vertical distance), highlighting its critical role in suppressing confident but geometrically invalid outliers. Together, these modules ensure robust convergence by effectively balancing positive guidance with noise rejection.

As shown in Table~\ref{tab:ablation_paramters}, the trigger threshold $\tau$ is most effective at $0.20$, while a looser $\tau = 0.40$ admits noisy inconsistent predictions and causes a 3.53\% performance drop. The soft label tolerance $\tau_{label}=0.20$ provides a robust neighborhood for pseudo-labels. Increasing $k$ from 1 to 10 improves average accuracy from 44.12\% to 46.47\%, with diminishing gains beyond $k=10$. Additional cross-dataset validation is provided in supplementary materials C.4.

\subsection{Qualitative Comparison}
To qualitatively evaluate the effectiveness of our framework, we compare the baseline Qwen3-VL-4B-Instruct~\cite{yang2025qwen3} with our proposed TTL-SR. As illustrated in Figure~\ref{examples}, the baseline frequently produces spatial predictions that are numerically imprecise or physically inconsistent, particularly in scenarios requiring complex geometric reasoning.
In contrast, predictions refined by TTL-SR exhibit markedly improved numerical precision and stronger adherence to geometric constraints. By leveraging geometric consistency as an intrinsic supervisory signal, our method effectively corrects implausible outputs, yielding predictions that are both closer to ground truth values and more coherent under physical laws.
Moreover, TTL-SR consistently improves performance across diverse scenes, suggesting that the refinement process generalizes beyond specific instances. These qualitative results indicate that our approach enhances not only the final predictions but also the underlying reasoning process, steering the model toward more physically grounded solutions.
Additional qualitative results, along with failure case analysis, are provided in supplementary materials C.2 and C.3.

\section{Conclusion}
In this paper, we propose \textbf{TTL-SR}, a geometry-aware test-time learning framework for improving quantitative spatial reasoning in VLMs. By leveraging geometric consistency as an intrinsic supervisory signal, TTL-SR enables unsupervised adaptation during test time, aligning model predictions with underlying spatial constraints and activating latent spatial knowledge.
Extensive experiments across multiple benchmarks demonstrate consistent improvements across both spatially specialized and general-purpose VLMs. Beyond quantitative gains, TTL-SR produces more physically plausible and logically coherent predictions, suggesting a shift toward more structured and reliable reasoning. More broadly, our findings suggest that incorporating explicit geometric structure into test-time learning can bridge data-driven prediction and physically grounded reasoning, with TTL-SR serving as a lightweight alternative to retraining. Future work will explore more flexible constraint formulations and extend the framework to a wider range of complex reasoning tasks and settings.

\begin{acks}
The research is partially supported by the National Natural Science Foundation of China (No. 62576139, 62176093) and the National Key Research and Development Program of China (2023YFC3502900).
\end{acks}







\balance
\bibliographystyle{ACM-Reference-Format}
\bibliography{sample-base}
\end{document}